\documentclass[runningheads]{llncs}
\usepackage{float}
\usepackage{graphicx}
\usepackage{subcaption} 
\usepackage{multirow}

\usepackage{amsmath,amssymb} 

\DeclareMathOperator*{\argmin}{arg\,min}

\usepackage{hyperref}
\usepackage{color}
\usepackage[normalem]{ulem}
\newcounter{ToDo}
\newcounter{gaocomm}

\definecolor{blue-violet}{rgb}{0.54, 0.17, 0.89}
\definecolor{mygreen}{rgb}{0.0, 0.5, 0.0}
\definecolor{awesome}{rgb}{1.0, 0.13, 0.32}
\definecolor{bostonuniversityred}{rgb}{0.8, 0.0, 0.0}

\usepackage{algorithm}
\usepackage[noend]{algpseudocode}
\usepackage{xpatch}
\makeatletter
\def\BState{\State\hskip-\ALG@thistlm}
\makeatother
\algrenewcommand\alglinenumber[1]{{\sffamily\footnotesize#1}}
\makeatletter
\xpatchcmd{\algorithmic}{\itemsep\z@}{\itemsep=1ex plus2pt}{}{}
\makeatother
\usepackage{amssymb}

\begin{document}
\title{Cluster Developing 1-Bit Matrix Completion}
%
%
\author{Chengkun Zhang \and
Junbin Gao \and Steven Lu}
\authorrunning{C. Zhang et al.}
%
\institute{University of Sydney, NSW, Australia  
\email{\{chengkun.zhang,junbin.gao,steven.lu\}@sydney.edu.au}}

\maketitle              
\begin{abstract}
Matrix completion has a long-time history of usage as the core technique of recommender systems. In particular, 1-bit matrix completion, which considers the prediction as a ``Recommended'' or ``Not Recommended'' question, has proved its significance and validity in the field. However, while customers and products aggregate into interacted clusters, state-of-the-art model-based 1-bit recommender systems do not take the consideration of grouping bias. To tackle the gap, this paper introduced Group-Specific 1-bit Matrix Completion (GS1MC) by first-time consolidating group-specific effects into 1-bit recommender systems under the low-rank latent variable framework. Additionally, to empower GS1MC even when grouping information is unobtainable, Cluster Developing Matrix Completion (CDMC) was proposed by integrating the sparse subspace clustering technique into GS1MC. Namely, CDMC allows clustering users/items and to leverage their group effects into matrix completion at the same time. Experiments on synthetic and real-world data show that GS1MC outperforms the current 1-bit matrix completion methods. Meanwhile, it is compelling that CDMC can successfully capture items' genre features only based on sparse binary user-item interactive data. Notably, GS1MC provides a new insight to incorporate and evaluate the efficacy of clustering methods while CDMC can be served as a new tool to explore unrevealed social behavior or market phenomenon.

\end{abstract}

\section{Introduction}
Recommender systems aim at improving customers' experience by maximizing the use of the available information, including $(i)$ user-item interactive data, such as ratings or clicking behavior, and $(ii)$ attribute information, such as category or context profiles. Methods that utilize the interaction data are referred as collaborative filtering \cite{deshpande2004item,linden2003amazon,resnick1994grouplens,sarwar2001item} while the other methods that use the textual information are referred as content-based methods \cite{billsus2000user,pazzani1997learning,shoham1997combining}. In particular, collaborative filtering is a method predicting the missing ratings given by a specific user to a specific item. Based on the idea that users and items are highly correlated to each other, the unspecified ratings can be estimated via learning the hidden relations. 
    
Collaborative filtering can be seen as a special case of matrix completion task. It has become a cornerstone of most powerful recommender systems while it is mainly founded on two main streams of methods: neighbourhood-based methods \cite{deshpande2004item,joaquin1999memory,linden2003amazon,resnick1994grouplens,sarwar2001item} and model-based methods \cite{breese1998empirical,bell2007modeling,koren2008factorization,paterek2007improving,takacs2008investigation,takacs2009scalable}. Though neighbourhood-based methods are easy to interpret and implement, they cannot extract enough information and suffer from low prediction accuracy when observed data is sparse. In this case, dimension reduction methods \cite{bell2007modeling,billsus1998learning,sarwar2000application} and graphs \cite{gori2007itemrank,pirotte2007random} were tried to address the sensitivity issue. Alternatively, model-based methods define a parameterized model which can be optimized by the available data during the training process. Numerous model-based approaches were tested in previous research, including Support Vector Machines \cite{grvcar2006knn}, Maximum Entropy \cite{zitnick2004maximum}, Boltzmann Machines \cite{salakhutdinov2007restricted} and Singular Value Decomposition (SVD) \cite{koren2008factorization,paterek2007improving,takacs2008investigation,takacs2009scalable}.
    
Under the assumption that the continuation of data points is convincing and compelling, standard collaborative filtering methods take observed entries of a rating matrix as real numbers. However, the adequacy of this measurement is undoubtedly questionable when intervals between data points are different. For instance, personal judgments from different customers vary as a result of personality. Say, generous customers tend to give fairly higher ratings than curmudgeon customers. Thus, instead of taking data as continuous numbers, it is more feasible considering them as categories, especially the binary case. For instance, researchers \cite{BhaskarJavanmard2015,cai2013max,davenport20141} use a small part of binary subset generated by the real-valued entries, namely `$+1$' for ``Recommended'' and `$-1$' for ``Not Recommended''. Experiments show their approaches perform significantly better than continuous matrix completion methods. 
    
Although 1-bit matrix completion has proven its success in recommender systems, same as most other matrix completion methods, it suffers from a fundamental limitation: every user/item is treated merely as standalone individuals, which arrogantly ignores the homogeneity of products and the clustering characteristic of social behaviors. For instance, some fundamental management theory points out that people have a propensity of conformity nature based on demographic, psychographic and behavioral variables \cite{kotler2009marketing}. Some recent research was noticed focusing on integrating preliminary clusters into continuous matrix completion task \cite{bi2017group} and experiments demonstrated that their approach outclassed traditional SVD methods. However, to the best of our knowledge, so far there is not any 1-bit matrix completion methods taking cluster information into consideration. Moreover, state-of-the-art recommender systems either take clustering as an independent task or treat clusters as preliminaries, there is not any existing method for summarizing clusters along with matrix completion. Since the clustering nature of individuals plays a vital role in social behavior research, it is consequently significant to introduce a new method that learns the clusters, on the other hand also to utilize the clustering effects for matrix completion. In this work, we focus on two tasks: (i) integrating group information into 1-bit matrix completion, namely \textit{group-specific 1-bit matrix completion} (GS1MC), and (ii) proposing an efficient algorithm of cluster developing matrix completion on the binary case, viz. \textit{cluster developing matrix completion} (CDMC). 
    
To exploit the grouping effects, based on current latent variable model, we expand the scope of quantized matrix completion to developing clusters automatically as well as leveraging their effects. The proposed methods can be used to take advantages of preliminary known user/item clusters or learn the groups during the training process according to the subspace correlations of targets. Experimentally, we show that the proposed GS1MC outperforms existing known model-based 1-bit matrix completion methods. And more importantly, CDMC successfully captures targets' generic features and achieves convergence of both user/item clusters.
    
The rest of the paper is constructed as follows: In Section \ref{Sec:2}, we discuss preliminary knowledge and background of the problem setting. In Section \ref{Sec:3}, we introduce \textit{group-specific 1-bit matrix completion}(GS1MC). In Section \ref{Sec:4}, the method is further extended to positively learn the cluster identities: \textit{cluster developing matrix completion}(CDMC). In Section \ref{Sec:5}, we evaluate our method on synthetic data as well as a real-world application. Section \ref{Sec:6} presents the discussion and future aspiration.

\section{Background}\label{Sec:2}
In this section, we discuss some preliminary knowledge of the research, including traditional SVD-based matrix completion, the framework of probabilistic 1-bit matrix completion and sparse subspace clustering techniques.

\subsection{Matrix Completion}\label{sec2.1}
Consider $\mathbf{\hat{R}} = (\hat{r}_{ui})_{n_1 \times n_2}$ as the original utility matrix, where $n_1$ and $n_2$ are the number of users and items, respectively. Within $\mathbf{\hat{R}}$, each $\mathbf{\hat{r}}_{ui}$ is the explicit feedback given by user $u$ towards item $i$ of a scale, e.g., from $1$ to $5$, where the intervals probably differ as a result of personal bias. Regularized SVD (RSVD) \cite{SimonFunk2006} predictor assumes $\mathbf{\hat{R}}$ as a low-rank matrix because of instance correlations and make the approximation (prediction) by:
\begin{equation}
            r_{ui} = \mathbf{u}_u \mathbf{v}_i^T,
\end{equation}
where $\mathbf{u}_u$ and $\mathbf{v}_i$ are $K$-dimensional latent variables associated to user $u$ and item $i$, respectively. RSVD estimates the latent variables by minimizing the sum of residuals of observed entries via gradient descent method with a regularization term:
        \begin{equation*}
            \hat{\bold{u}}_u = \argmin_{\bold{u}_u} \sum_{i \in \Omega_u} (r_{ui} - \bold{u}_u\bold{v}_i^T)^2 + \lambda \|\bold{v}_i\|^2_2
        \end{equation*}
        and 
        \begin{equation*}
            \hat{\bold{v}}_i = \argmin_{\bold{v}_i} \sum_{u \in \Omega_i} (r_{ui} - \bold{u}_u\bold{v}_i^T)^2 + \lambda \|\bold{u}_u\|^2_2,
        \end{equation*}
        where $\Omega_u$ denotes all items rated by user $u$ and $\Omega_i$ stands for all users who rated item $i$. 
        
        As the most fundamental SVD method, RSVD has been extended in different directions. For instance, a variety of regularization terms were applied for specific considerations \cite{zhu2016personalized}, and biased version of SVD methods \cite{koren2015advances,koren2009matrix,paterek2007improving} were also introduced. To take the advantages of the general preference of each user and discrimination of each item, a set of biasing variables were incorporated in biased SVD methods. Then, apart from taking individual-specific bias, users/items can also be allocated into clusters and aggregated with group effects. For instance, taking preliminary cluster identities as inputs, a set of latent variables representing the group bias \cite{bi2017group} can be learned via the training process. 
        
    \subsection{1-Bit Matrix Completion}\label{sec2.2}
        Though matrix completion methods have been used for recommender systems for long, 1-bit matrix completion \cite{davenport20141} has been officially introduced lately. Varied from the continuous model which applies numerical computation on discrete rating data directly, original observation is converted into a binary matrix $\mathbf{\hat{Y}}$ by comparing each observed entry to the average rating score. Then, the objective of the task is formalized as learning an $n_1 \times n_2$ latent variable matrix $\mathbf{M}$. The predicted binary feedback is finally computed by:
        \begin{equation}
            Y_{ui}=
                \begin{cases}
                +1, &\mbox{with probability }f(M_{ui})  \\
                -1, &\mbox{with probability }1-f(M_{ui})
                \end{cases}
            \mbox{ $(u,i) \in \Omega$,}
            \label{1-bit}
        \end{equation}   
        where $\Omega$ is the set of all the observed entries and $f$ can be the Sigmoid function defined as: 
        \begin{align}
            f(z) = \frac1{1+\exp\{-z\}}. \label{Eq2a}
        \end{align}
        
        Similar to other low-rank matrix completion methods, a wide variety of approaches have been applied to constrain the latent variable matrix. For instance, a trace-norm \cite{davenport20141} was considered under the assumption of uniform sampling. Then, a max-norm method as a convex relaxation \cite{cai2013max} was explored under a general sampling model.  Moreover, the theory has been extended further to discuss the exact low-rank constraint \cite{BhaskarJavanmard2015}. However, all these existing 1-bit matrix completion methods treat every instance as autonomous individuals. In other words, predictions have been made generously, ignoring the ground truth that users/items tend to have a specific baseline or belong to certain clusters. Furthermore, as far as we know, there is not any methodology that can both learn the cluster identities and leverage their group effects for matrix completion at the same time. 
        
    \subsection{Sparse Subspace Clustering}\label{sec2.3}
        Sparse subspace clustering (SSC) \cite{elhamifar2013sparse} aims at clustering data points in their low-dimensional subspace via the \textit{self expressive matrix}, which represents each instance by an affine combination of other points within the same subspace. 
        
        Nevertheless, in terms of the fact that representations for each data point by the other should be as sparse as possible, which results in an NP-hard problem, a convex relaxation must be proposed to get around the NP difficulty. Thus, SSC formalizes the original problem as a $l_1$-norm optimization task. Take the most standard procedure as an example, SSC assumes the whole noise-free dataset $\mathbf{X} \in \mathbb{R}^{D \times N_l}$ can be separated into $n$ subspaces $\{S_l\}^n_{l=1}$ of dimensions $D = \{d_l\}^n_{l=1}$. Alternatively speaking, the matrix of the whole dataset can be written as:
        \begin{equation*}
            \mathbf{X} \triangleq [\mathbf{x}_1 ... \mathbf{x}_N] = [\mathbf{X}_1 ... \mathbf{X}_n]\Upsilon,
        \end{equation*}
        where $\Upsilon$ is an unknown permutation matrix and $\mathbf{X}_l \in \mathbb{R}^{D \times N_l}$ is a subset of the data points lying in $\mathbf{S}_l$, namely a $d_l$-rank matrix of $N_l$ points ($N_l > d_l$). Now, each data point $\mathbf{x}_i \in \cup^n_{l=1}S_l$ can be reconstructed by a combination of other points within the same subspace as:
        \begin{equation}
            \mathbf{x}_i = \mathbf{X}\mathbf{c}_i, c_{ii} = 0. \label{nphard}
        \end{equation} 
        Then, different norm functions can be applied for the estimation of \eqref{nphard}. 
        Finally the problem is defined as, under the  $l_1$-norm constraint,
        \begin{equation}
            \min \|\mathbf{C}\|_1 \text{ s.t. }\ \mathbf{X} = \mathbf{XC}, \ \text{diag}(\mathbf{C}) = 0,
            \label{sef-expressive}
        \end{equation}
        where $\mathbf{C} \triangleq [\mathbf{c}_1 \mathbf{c}_2 ... \mathbf{c}_N] \in \mathbb{R}^{N \times N}$ corresponds to the non-trivial subspace-sparse representation for all the data points $\mathbf{x}_i$s.
        
        Since user-item interaction data is exceedingly sparse and high-dimensional, many dimensions are irrelevant and covered by noise. In the meantime, the correlation between individuals can be interpreted as similarities of their private latent variable, which is not strictly around any centroids. Thus, conventional clustering methods that utilizing the spatial proximity is not applicable in this case. Differently, subspace clustering methods aim at grouping the points that are not necessarily close but lie in the same subspace, which does not depend on the spatial characteristic of the data. Moreover, as sparse subspace clustering deploys a convex approach to pick out the sparse representation of each point, the optimization process automatically eliminates some common issues of clustering methods, such as sensitivity to the ideal cluster size and bordering matter of the overlapped subspace. 
        
\section{Group-specific 1-Bit Matrix Completion (GS1MC)}\label{Sec:3}
    In this section, we integrate group effects into 1-bit matrix completion task such that biases of clusters can be learned along with latent variable training process.
     
    \subsection{Model Framework}
        Suppose $\mathbf{\hat{Y}}$ is the observed $n_1 \times n_2$ binary rating matrix with entries equal to `$+1$' or `$-1$', corresponding to ``interested'' or ``not interested'', where $n_1/n_2$ is the number of users/items, the ``not observed'' entries are represented by `$0$'. $\Omega$ stands for the observed user-item pairs, i.e. entries with same indexes as `$+1$' and `$-1$' in $\mathbf{\hat{Y}}$.  We construct the latent variable matrix as $\mathbf{M} \in \mathbb{R}^{n_1 \times n_2}$. To make predictions for missing entries by \eqref{1-bit}, our main objective is to find the estimation of $\mathbf{M}$ that best explains the observed data.
        
        Since it has been proved that the exact low-rank method results in a high convergence rate \cite{BhaskarJavanmard2015}, especially when the fraction of revealed entries is small (\textit{cold-start problem}), we choose to apply an exact low-rank constraint on $\mathbf{M}$. We assume that every user/item is classified into one single user/item group, respectively. We formulate the latent variable matrix $\mathbf{M}$ by integrating group bias into matrix factorization. Then each entry in $\mathbf{M}$ can be written as: 
            \begin{equation}
                    M_{ui} = (\mathbf{p}_u + \mathbf{s}_{v_u})'(\mathbf{q}_i + \mathbf{t}_{j_i}). \label{m_factorization}
            \end{equation}
        Here $\mathbf{p}_u \in \mathbb{R}^K$ and $\mathbf{q}_i \in \mathbb{R}^K$ are $K$-dimensional latent factors standing for user \textit{u}'s preference and item \textit{i}'s character, while $\mathbf{s}_{v_u} \in \mathbb{R}^K$ and $\mathbf t_{j_i} \in \mathbb{R}^K$ represent biases of clusters that  individuals belong to. For instance, $\mathbf{s}_{v_u}$ means the cluster effect of user cluster $v_u$, i.e. the cluster user $u$ belongs to. Here we have assumed that there are $m_1$ users clusters and $m_2$ item clusters, such that $v_u \in \{1, 2, ..., m_1\}$ and $j_i \in \{1, 2, ..., m_2\}$. Then, the group effects of the user and item clusters can be formalized as:
            \[
                \mathbf S_U = [\mathbf s_1, \mathbf s_2, ..., \mathbf s_{m_1}]^T   \in \mathbb{R}^{m_1\times K} 
                 \quad \textrm{and} \quad 
                \mathbf T_J = [\mathbf t_1, \mathbf t_2, ..., \mathbf t_{m_2}]^T   \in \mathbb{R}^{m_2\times K}.
            \]
        For the sake of convenience, in terms of matrix notations, we assume the user-item interaction data $\mathbf{\hat{Y}} = (\hat{Y}_{ui})$ and its corresponding latent variable $\mathbf M = (M_{ui})$ have been permuted such that the first $U_1$ rows corresponds user cluster 1, followed by $U_2$ rows corresponding to user cluster 2, ..., and the last $U_{m_1}$ rows corresponding to user cluster $m_1$. Similarly, the columns have been rearranged accordingly. After this alteration, the decomposition \eqref{m_factorization} can be written as the following matrix format:
            \begin{equation}
                \mathbf M = (\mathbf P + \mathbf S)(\mathbf Q +\mathbf T)^T 
                \label{M_factorization_ST}
            \end{equation}
        where 
            \begin{align*}
                &\mathbf P = [\mathbf{p}_1, \mathbf{p}_2, ..., \mathbf{p}_{n_1}]^T \in\mathbb{R}^{n_1\times K};   \\
                &\mathbf Q = [\mathbf{q}_1, \mathbf{q}_2, ..., \mathbf{q}_{n_2}]^T \in\mathbb{R}^{n_2\times K}; \\
                &\mathbf S = [\mathbf s_1\mathbf 1^T_{U_1}, \mathbf s_2\mathbf 1^T_{U_2}, ..., \mathbf s_{m_1}\mathbf 1^T_{U_{m_1}}]^T \in\mathbb{R}^{n_1\times K};\\
                &\mathbf T = [\mathbf t_1\mathbf 1^T_{J_1}, \mathbf t_2\mathbf 1^T_{J_2}, ..., \mathbf t_{m_2}\mathbf 1^T_{J_{m_2}}]^T\in\mathbb{R}^{n_2\times K}. 
            \end{align*}
        Here $\mathbf 1_{m}$ stands for $m$-dimensional (column) vector of all `1's. In other words, instances of group effects matrix $\mathbf{S}_U$ and $\mathbf{T}_J$ have been duplicated in order to match the dimension of matrix $\mathbf{P}$ and $\mathbf{T}$. For the convenience of the transformation between $\mathbf{S}$, $\mathbf{T}$ and $\mathbf{S}_U$, $\mathbf{T}_J$, we define the following two matrices:
            \[
            \mathbf{I}_U^{m_1\times n_1} = \begin{bmatrix}\mathbf 1^T_{U_1} & \mathbf 0 & \cdots & \mathbf 0\\
            \mathbf 0 & \mathbf 1^T_{U_2} & \cdots & \mathbf 0 \\
            \vdots & \vdots & \ddots & \vdots\\
            \mathbf 0&\mathbf 0 & \cdots & \mathbf 1^T_{U_{m_1}}\end{bmatrix} 
             \quad \textrm{and} \quad 
            \mathbf{I}_J^{m_2\times n_2} = \begin{bmatrix}{\mathbf 1}^T_{J_1} & \mathbf 0 & \cdots & \mathbf 0\\
            \mathbf 0 & \mathbf 1^T_{J_2} & \cdots & \mathbf 0 \\
            \vdots & \vdots & \ddots & \vdots\\
            \mathbf 0&\mathbf 0 & \cdots & \mathbf 1^T_{J_{m_2}}\end{bmatrix}.
            \]
        Thus, it is clear that $\mathbf{S}$, $\mathbf{T}$ and $\mathbf{S}_U$, $\mathbf{T}_J$ can be transformed to each other by:
        \begin{equation}
            \mathbf S = \mathbf{I}_U^T  \mathbf{S}_U
            \text{ and }
            \mathbf{T} = \mathbf{I}_J^T  \mathbf{T}_J.
            \label{ST_SuTj}
        \end{equation}
        Then, \eqref{M_factorization_ST} can be rewritten as:
        \begin{equation}
            \mathbf{M} = (\mathbf{P}+\mathbf{I}^T_U\mathbf{S}_U)(\mathbf{Q}+\mathbf{I}^T_J\mathbf{T}_J).
            \label{M_factorization}
        \end{equation}
        
    \subsection{Objective and Optimization}
        
        Following the objective function of basic 1-bit matrix completion method \cite{BhaskarJavanmard2015}, the fundamental loss function is defined as:
        \begin{equation*}
                 F_{\Omega,\mathbf{\hat{Y}}}(\mathbf M)=-\sum_{(u,i)\in\Omega} \{\mathbb{I}_{(\hat{Y}_{ui}=1)}\log(f(M_{ui})) \\
                    + \mathbb{I}_{(\hat{Y}_{ui}=-1)}\log(\mathbf{1}-f(M_{ui}))\},
            \label{F}
        \end{equation*}
        where $f(\mathbf M)$ is the matrix operation of applying $f$ over $\mathbf M$ element-wise, and $\boldsymbol{1}$ is the all 1's matrix. Here $\mathbb{I}_\mu$ is the indicator function, i.e. $\mathbb{I}_\mu = 1$ when $\mu$ is true, else $\mathbb{I}_\mu = 0$. $\mathbb{I}_\mu$ can be implemented as two mask matrices $\mathbf{Y}_1^{n1 \times n2} = (Y_1(u,i))$  and $\mathbf{Y}_{-1}^{n1 \times n2} = (Y_{-1}(u,i))$ of the same size as $\mathbf{M}$, where $Y_1(u,i) = 1$ if $\hat{Y}_{ui}=1$, otherwise $Y_1(u,i) = 0$, and $Y_{-1}(u,i) = 1$ if $\hat{Y}_{ui}=-1$, otherwise $Y_{-1}(u,i) = 0$. Then, the fundamental loss function can be transformed into:
            \begin{equation}
                    F_{\Omega,\mathbf{\hat{Y}}}(\mathbf M) = -\sum_{\Omega}( Y_1\circ \log f(\mathbf M) \\
                         + Y_{-1}\circ \log (\boldsymbol{1} - f(\mathbf M)),
            \end{equation} 
        where $\circ$ means the element-wise product of two matrices. We notate $\Gamma = (\mathbf{P,Q,S_U,T_J})$ and $R^0 = \{\hat{Y}_{ij}:(i,j) \in \Omega\}$. After adding the regularization term, the new loss function can be formulated as:
            \begin{equation}
                    L(\Gamma|R^0)= F_{\Omega,\mathbf{\hat{Y}}}(M)
                        + \lambda(\|\mathbf{P}\|^2_F + \|\mathbf{S}_U\|^2_F 
                        + \|\mathbf{Q}\|^2_F + \|\mathbf{T}_J\|^2_F).
            \end{equation}
        Our goal is to predict the missing entries of the rating matrix, which can be computed by:
            \begin{equation}
                \hat{\Gamma} = \argmin_\Gamma L(\Gamma | R^0).
                \label{loss_function}
            \end{equation}
        We solve the optimization problem \eqref{loss_function} via the Alternating direction method of multipliers (ADMM). Firstly, to update the latent factors of users and user clusters, we fix $\mathbf{Q}$ and $\mathbf{T}_J$, and minimize \eqref{loss_function} by estimating $\mathbf{\hat{P}}$ and $\mathbf{\hat{S_U}}$:
        
        \begin{equation}
            \mathbf{\hat{P}} = \argmin_\mathbf{P} 
                        + \lambda\|\mathbf{P}\|^2_F \}, \label{Gao:1}
        \end{equation}
        \begin{equation}
            \mathbf{\hat{S}}_U = \argmin_{\mathbf{S}_U} 
                        + \lambda\|\mathbf{S}_U\|^2_F \}.\label{Gao:2}
        \end{equation}



        Then for items and item clusters, we fix $\mathbf{P}$ and $\mathbf{S}_U$, conducting following computations:
        \begin{equation}
            \mathbf{\hat{Q}} = \argmin_\mathbf{Q} 
                        + \lambda\|\mathbf{Q}\|^2_F \}, \label{Gao:3}
        \end{equation}
        \begin{equation}
            \mathbf{\hat{T}}_J = \argmin_{\mathbf{T}_J} 
                        + \lambda\|\mathbf{T}_J\|^2_F \}. \label{Gao:4}
        \end{equation}
        
        Each of sub-problems \eqref{Gao:1} - \eqref{Gao:4} can be solved by the gradient descent algorithm. We can work out the gradient in the following way. First we take $f$ as the Sigmoid function defined in \eqref{Eq2a}, then it is easy to check that: 
            \[
            \frac{\partial F_{\Omega,\mathbf{\hat{Y}}}(\mathbf M)}{\partial \mathbf M} = Y_1 \circ (f(\mathbf M) - \boldsymbol{1}) + Y_{-1}\circ f(\mathbf M).
            \]
        Considering \eqref{M_factorization_ST}, with the matrix differentiation chain rule, it can be proved that:
            \begin{align}
            \frac{\partial F_{\Omega,\mathbf{\hat{Y}}}(\mathbf M)}{\partial \mathbf P} & = [Y_{-1} + (Y_1+Y_{-1})\circ (f(\mathbf M) - \boldsymbol{1})](\mathbf Q + \mathbf T)\\
            \frac{\partial F_{\Omega,\mathbf{\hat{Y}}}(\mathbf M)}{\partial \mathbf Q} & = [Y^T_{-1} + (Y^T_1+Y^T_{-1})\circ (f(\mathbf M^T) - \boldsymbol{1})]^T(\mathbf P + \mathbf S).
            \end{align} 
        On the one hand, we have
            \[
            \frac{\partial F_{\Omega,\mathbf{\hat{Y}}}(\mathbf M)}{\partial \mathbf S} = \frac{\partial F_{\Omega,\mathbf{\hat{Y}}}(\mathbf M)}{\partial \mathbf P}
             \quad \textrm{and} \quad
            \frac{\partial F_{\Omega,\mathbf{\hat{Y}}}(\mathbf M)}{\partial \mathbf T} = \frac{\partial F_{\Omega,\mathbf{\hat{Y}}}(\mathbf M)}{\partial \mathbf Q}.
            \]
        On the other hand, according to \eqref{ST_SuTj}, it is clear to state that:  
            \[
            \frac{\partial \mathbf S}{\partial \mathbf S_U} = {I_K\otimes I^T_U}   \quad \textrm{and} \quad
            \frac{\partial \mathbf T}{\partial \mathbf T_J} = {I_K\otimes I^T_J}. 
            \]
        According to the chain rules, we finally get:
            \[
            \frac{\partial F_{\Omega,\mathbf{\hat{Y}}}(\mathbf M)}{\partial \mathbf S_U} = I_U \frac{\partial F_{\Omega,\mathbf{\hat{Y}}}(\mathbf M)}{\partial \mathbf S}
              \quad \textrm{and} \quad
            \frac{\partial F_{\Omega,\mathbf{\hat{Y}}}(\mathbf M)}{\partial \mathbf T_J} = I_J \frac{\partial F_{\Omega,\mathbf{\hat{Y}}}(\mathbf M)}{\partial \mathbf T}.
            \]
        In other words, the sum of the first $U_1$ rows of $\frac{\partial F_{\Omega,\mathbf{\hat{Y}}}(\mathbf M)}{\partial \mathbf S}$ is the first row of $\frac{\partial F_{\Omega,\mathbf{\hat{Y}}}(\mathbf M)}{\partial \mathbf S_U}$, the sum of the next $U_2$ rows of $\frac{\partial F_{\Omega,\mathbf{\hat{Y}}}(\mathbf M)}{\partial \mathbf S}$ is the second row of $\frac{\partial F_{\Omega,\mathbf{\hat{Y}}}(\mathbf M)}{\partial \mathbf S_U}$, ..., and the sum of the last $U_{m_1}$ rows of  $\frac{\partial F_{\Omega,\mathbf{\hat{Y}}}(\mathbf M)}{\partial \mathbf S}$ becomes the $m_1$-th row (the last row) of $\frac{\partial F_{\Omega,\mathbf{\hat{Y}}}(\mathbf M)}{\partial \mathbf S_U}$.  The similar way can be used to construct $\frac{\partial F_{\Omega,\mathbf{\hat{Y}}}(\mathbf M)}{\partial \mathbf T_J}$ from $\frac{\partial F_{\Omega,\mathbf{\hat{Y}}}(\mathbf M)}{\partial \mathbf T}$.

\section{Cluster Developing Matrix Completion (CDMC)}\label{Sec:4}
    In this section, we intend to learn the cluster identities of users/items during the latent variable training process and integrate the clustering results with group-specific matrix completion. 
    
    \subsubsection{Problem Setting}    
    The model (GS1MC) proposed in Section \ref{Sec:3} takes cluster identities as preliminary information. However, in most practical scenarios, it might be inaccessible to such details, especially for the \textit{cold-start problem}. Secondly, since the original binary user-item interaction data is extremely sparse, it is controversial to apply standard clustering techniques on it directly. Moreover, common clustering methods may take advantage of distance between points to divide the space into different partitions. Nevertheless, regarding a latent variable model, market segments may not necessarily congregate based on spatial proximity but lie in a subspace. Thus, found on GS1MC, we aim at clustering users/items that belong to a union of low-dimensional subspace respectively.
    
    A common dilemma for most clustering techniques is the drawback that they might be decidedly sensitive to improper initialization, such as cluster size and centroids. As long as the size of user/item clusters is unrevealed and each data points can have an infinite number of expressions in terms of the other, we incorporate sparse subspace clustering (SSC) technique to optimize a sparse representation among these expressions through a convex realization approach. 
    
    \subsubsection{Algorithm}
    Based on GS1MC, we extend the scope of the method to developing clusters during the latent variable training process. 
    
    In the last session, we deploy ADMM to optimize latent variables $\mathbf{P},\mathbf{S}_U,\mathbf{Q}$ and $\mathbf{T}_J$ in an iterative manner. Now, to develop clusters based on the gradually recovering matrix, after each iteration of updating latent variables $\mathbf{P},\mathbf{S}_U,\mathbf{Q}$ and $\mathbf{T}_J$, we construct the rating likelihood matrix $f(\mathbf{M})$ and $f(\mathbf{M})^T$ via \eqref{M_factorization} and \eqref{Eq2a}. We consider the rating likelihood matrix $f(\mathbf{M})$ lies in $m_2$ disjoint subspaces $\{S_i\}^{m_2}_{i = 1}$ while $f(\mathbf{M})^T$ lies in $\{S_i\}^{m_1}_{i = 1}$. According to Theorems 2 and 3 from \cite{elhamifar2013sparse}, we employ the $l_1$-norm relaxation of the \textit{self-expressive matrix} to obtain the sparse representation $\mathbf{C}_1$/$\mathbf{C}_2$ for users/items' features respectively, namely:
    \begin{align}
     \min \|\mathbf{C}_l\|_1 \text{ s.t. }\left\{
    \begin{aligned}
           &f(\mathbf{M}) = f(\mathbf{M})\mathbf{C}_1 \text{ or }\\ &f(\mathbf{M}) = \mathbf{C}^T_2f(\mathbf{M}),
\end{aligned} \right\}\;\; \text{diag}(\mathbf{C}_l) = 0,\;\; l = \{1,2\}.           
    \end{align}
    Here, each column of $\mathbf{C}_1$ and $\mathbf{C}_2$ stands for an user/item's hidden profile, and within each column, non-zero entries correspond to the other homogeneous points that lie in the same subspace with this point in the ideal case.
    
    Next, a non-directional weighted graph of $\mathbf{C}_1$ is built as $\mathbb{G_1} = (\mathbf{N_1}, \mathbf{W_1})$, where $\mathbf{N_1}$ is the nodes regarding all sparse representations in $\mathbf{C}_1$, and $\mathbf{W_1}$ is the weighted edges between each pair of $\mathbf{N_1}$. A natural choice of the weighted matrix is that the nodes within the same subspace will share non-zero weighted edges while the other edges are zero-weighted. Alternatively speaking, an affinity matrix can be constructed by $\mathbf{W}_1 = |\mathbf{C}_1| + |\mathbf{C}_1^T|$, where the non-zero entries represents latent variable pairs that actually lie in the same subspace. Then, we apply spectral clustering method on $\mathbf{W}_1$ to procure item clusters. Similar method is conducted to build $\mathbf{W_2}$ for user cluster developing. 
    
    After observing new clusters from last step, we update group identities of each user/item. Then, to leverage group effects of the latest clusters into matrix completion, we estimate latent variables $\mathbf{P},\mathbf{S}_U,\mathbf{Q},\mathbf{T}_J$ by \eqref{Gao:1} to \eqref{Gao:4} again. Thus, CDMC conducts sparse subspace clustering and GS1MC iteratively. The complete algorithm is shown in Algorithm \ref{algorithm_1}.

    \begin{algorithm}
    \caption{Cluster Developing 1-bit Matrix Completion}\label{euclid}
        \begin{algorithmic}[1]
            \Procedure{CDMC}{}
                \State Randomly initialize user/item groups
                \State Update latent variables $\mathbf{P},\mathbf{S}_U,\mathbf{Q},\mathbf{T}_J$ by \eqref{Gao:1} to \eqref{Gao:4}
              \BState \emph{loop}:
                \State Construct $f(\mathbf{M})$ Matrix by \eqref{M_factorization} and \eqref{Eq2a}
                \State Build adjacency matrices $\mathbf{C}_1$, $\mathbf{C}_2$, weighted graphs $\mathbf{W}_1$ and $\mathbf{W}_2$
                \State Apply spectral clustering on $\mathbf{W}_1$ and $\mathbf{W}_2$
                \State Update cluster identities
                \State Update latent variables $\mathbf{P},\mathbf{S}_U,\mathbf{Q},\mathbf{T}_J$ by \eqref{Gao:1} to \eqref{Gao:4} in a smaller inner loop
                \State If not converged, \textbf{goto} Step 4: \emph{loop}.
            \EndProcedure
        \end{algorithmic}
        \label{algorithm_1}
    \end{algorithm}

\section{Experiments}\label{Sec:5}
    In this section, we evaluate the proposed GS1MC and its extension CDMC, separately. The experiments are based on simulation analysis as well as benchmark comparison on a real-world dataset.
    
    \subsection{Dataset and Experiment Settings}
        To start with, to verify the effectiveness of GS1MC, a synthetic dataset with group information was designed in the following way. Firstly, we set $n_1 = 200$, $n_2 = 800$, $m_1 = 10$ and $m_2 = 10$. Then we generate $\mathbf{\hat{P}} \sim N(0,\mathbf{I}_{n_1 \times K})$ and $ \mathbf{\hat{Q}} \sim N(0,\mathbf{I}_{n_2 \times K})$, where $\mathbf{I}_{m \times K}$ is an $(m \times K)\times (m \times K)$-order identity matrix. To include the group information, we design $\mathbf{\hat{S}}_U = (\mathbf{\hat{s}}_v)$ and $\mathbf{\hat{T}}_J = (\mathbf{\hat{t}}_j)$, where $\mathbf{\hat{s}}_v \sim N(-2+0.4v, \mathbf{1}_K)$, $\mathbf{\hat{t}}_j \sim N(-3+0.6j, \mathbf{1}_K)$, $v \in \{1,...,m_1\}$ and $j \in \{1,...,m_2\}$. Then, we construct the latent variable matrix by $\mathbf{\hat{M}} = (\mathbf{\hat{P}}+\mathbf{I}^T_U\mathbf{\hat{S}}_U)(\mathbf{\hat{Q}}+\mathbf{I}^T_J\mathbf{\hat{T}}_J)$ and scale it so that $\|\mathbf{\hat{M}}\|_\infty = 1$. Now, we take the 1-bit transformation and add the noise by $f(\mathbf{\hat{M}}) + N(0,\mathbf{I}_{n_1 \times n_2})$. We keep a certain percentage $\pi$ of entries as observations, where $\pi$ is the observation rate.
        
        Notably, we also tested our methods on one of the most common recommender system benchmark dataset: Movielens $100k$ \cite{harper2016movielens}. This user-item interaction data consists of 100,000 ratings (1-5) from 943 users on 1682 movies. Following the problem settng of previous literature \cite{BhaskarJavanmard2015,cai2013max,davenport20141}, the original observations, scaled from 1 to 5, have been quantized as `+1' and `-1' according to whether they are above or below the average score. 
        
        The proposed method is implemented and tested in Matlab R2017b on a PC with Intel(R) Core(TM) i5-7600 CPU @ 3.500GHz and 8.00GB RAM.
    \subsection{Experiments on GS1MC}
        \subsubsection{Simulation Analysis}
        We set $K = 3$ and $K = 6$. Then we randomly split the data in terms of different training size, namely $\pi = \{25\%, 20\%,15\%,10\%\}$ for cross-validation. we assume the right group identities are preliminary and compare our method with a Trace-Norm approach \cite{davenport20141}. The tuning parameter $\lambda$ for the proposed method is selected as 37 by minimizing the average relative error while the parameters search for the Trace-Norm approach is embedded in the original implementation. The best results for both methods, shown in Table \ref{Table_RMSE_gsm}, are chosen among 100 replications. It is indicated that GS1MC has much smaller relative error compared to traditional 1-bit matrix completion, especially when the observed data is sparse (\textit{cold-start problem}) or when the latent variable have higher dimensions. 
        
        It is straightforward to comprehend the result: since group effects can be regarded as extra information compared to the observed sparse matrix, GS1MC can have a much more robust performance compared to the fundamental 1-bit matrix completion when the observed information is limited or when the complexity of the latent variable is high.
            \begin{table}
                \centering
                \begin{tabular}{|lcl|lcllcccc|l|}
                \hline
                 & No. of latent factors  &  &  & Observation Rate: $\pi$    &  &  & 10\% & 15\% & 20\% & \multicolumn{2}{c|}{25\%} \\ \cline{4-12} 
                 & \multirow{2}{*}{K = 3} &  &  & The Proposed Method &  &  & 1.00 & 0.85 & 0.78 & \multicolumn{2}{c|}{0.73} \\
                 &                        &  &  & Trace-Norm          &  &  & 1.89 & 1.74 & 1.67 & \multicolumn{2}{c|}{1.59} \\
                 & \multirow{2}{*}{K = 6} &  &  & The Proposed Method &  &  & 1.00 & 0.92 & 0.81 & \multicolumn{2}{c|}{0.74} \\
                 &                        &  &  & Trace-Norm          &  &  & 2.53 & 2.27 & 2.15 & \multicolumn{2}{c|}{2.02} \\ \hline
                \end{tabular}
                \caption{The relative error is computed by: $\frac{\|\mathbf{M} - \mathbf{\hat{M}}\|^2_F}{\|\mathbf{\hat{M}}\|^2_F}$. We have compared the results for synthetic data of different ranks and observation rates, namely $K=3$, $K=6$ and $\pi = \{25\%, 20\%,15\%,10\%\}.$}
                \vspace{-10mm}
                \label{Table_RMSE_gsm}
            
            \end{table}
            
            \begin{table}[]
                \centering
                \begin{tabular}{|c|c|c|c|c|c|c|}
                \hline
                 & \multicolumn{6}{c|}{\% Prediction accuracy} \\ \hline
                \% Training size & The proposed method & Exact-rank & HL & Logit & Trace-norm & Max-norm \\ \hline
                95 & 74.1 & 73.0 & 72.0 & 68.0 & 73.0 & 72.2 \\ \hline
                10 & 66.3 & 61.0 & - & - & 59.0 & 59.0 \\ \hline
                5 & 63.3 & 54.5 & - & - & 49.9 & 50.5 \\ \hline
                \end{tabular}
                \caption{Prediction accuracy of GS1MC versus methods in previous literature when the observation rate is 95\%, 10\% and 5\%.}
                \vspace{-10mm}
                \label{table_movielens_group}
            \end{table}
        \subsubsection{Movielens Dataset}
        Since the fact that the cluster information of most user-item interaction data is not available, to provide GS1MC cluster information, we group the original dataset according to their implicit feedback. Implicit feedback refers to the density of items receiving comments or the frequency of people giving feedback. In other words, people tend not to choose items randomly but choose things they already expected \cite{devooght2015dynamic}. Thus, implicit feedback only concerns about the identity of ratings irrespective of actual rating values. It is expected that people giving more ratings tend to be more curmudgeon while items with more feedback tend to have higher average ratings \cite{bi2017group}. Thus, we group users and items according to the number of ratings they have given or received. 
        

        We compared GS1MC with the other existing 1-bit matrix completion methods, namely: a) hinge loss with variational approximation (HL) \cite{cottet20161}, (b) Bayesian logistic model with variational approximation (Logit) \cite{cottet20161}, (c) the trace-norm frequentist logistic model (Trace-norm) \cite{davenport20141}, (d) the exact low-rank model (Exact-rank) \cite{BhaskarJavanmard2015} and (e) a max-norm constrained minimization approach (Max-norm) \cite{cai2013max}. Following their experiment setup, Movielens $100k$ dataset has been split into different training-test size (Note: Here, the training size is not the observation rate in the simulation anaysis). Since some methods are not open-sourced, we compared our results with the best results appeared in previous literature. The converged accuracy results are displayed in Table \ref{table_movielens_group}. It is easy to reveal that the proposed method has outperformed all the other baselines. Conspicuously, regarding the scenario when the training size is extremely small (5\%), our method has greatly boosted traditional binary matrix completion method by utilizing the group information.

    \subsection{Experiments on CDMC}
        \subsubsection{Robustness Analysis}
        As far as our knowledge, there is not any comparable baseline for clustering problems in recommender systems research. Thus, to evaluate the convergence performance of CDMC, we conduct the first experiment on Movielens $100k$ dataset.

\begin{figure}[H]
            \captionsetup[subfigure]{justification=centering}
            \centering
                \begin{subfigure}[b]{0.46\textwidth}
                    \includegraphics[width=\textwidth]{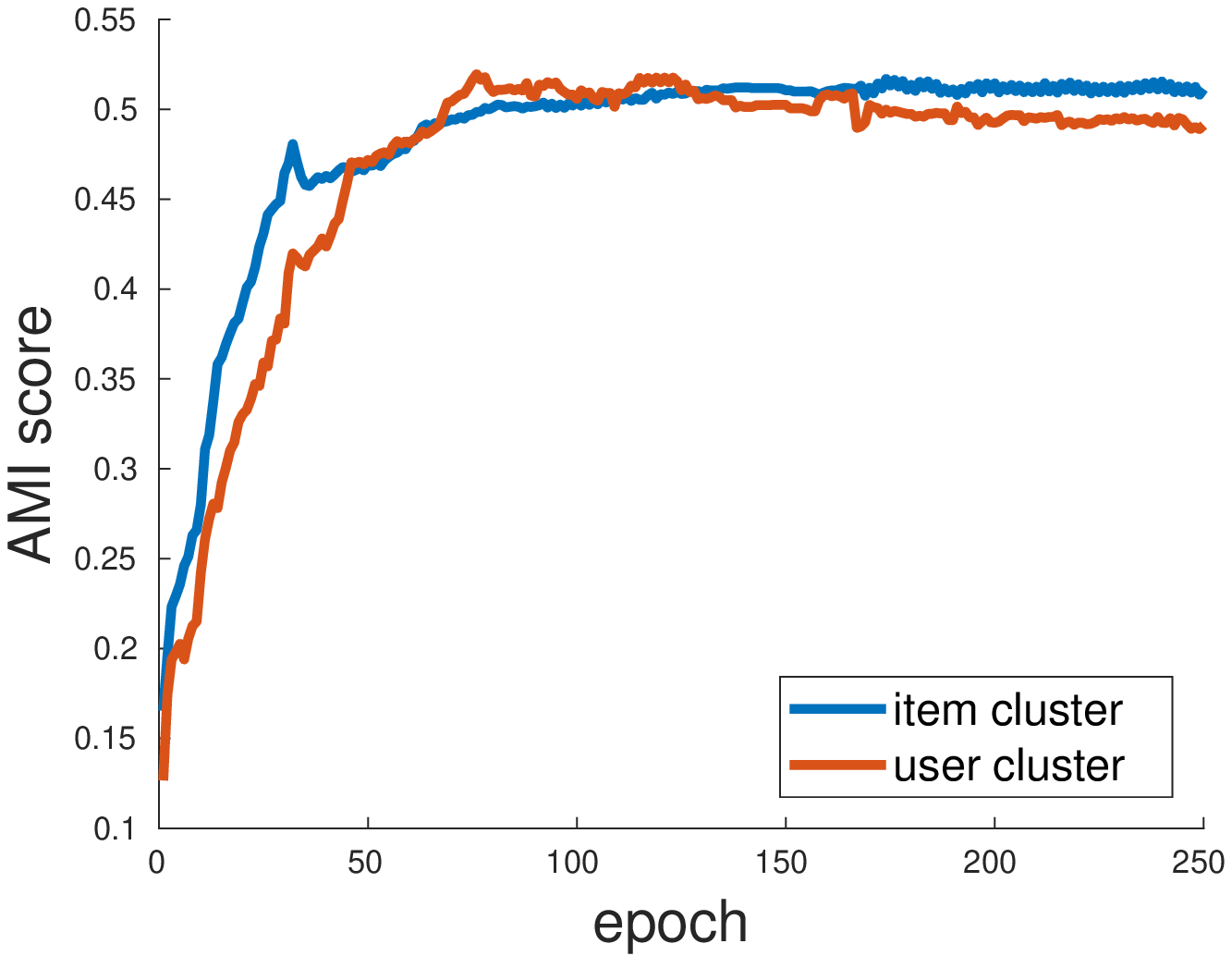}
                    \caption{Cluster convergence.}
                    \label{AMI_cluster}
              \end{subfigure}
              \begin{subfigure}[b]{0.46\textwidth}
                    \includegraphics[width=\textwidth]{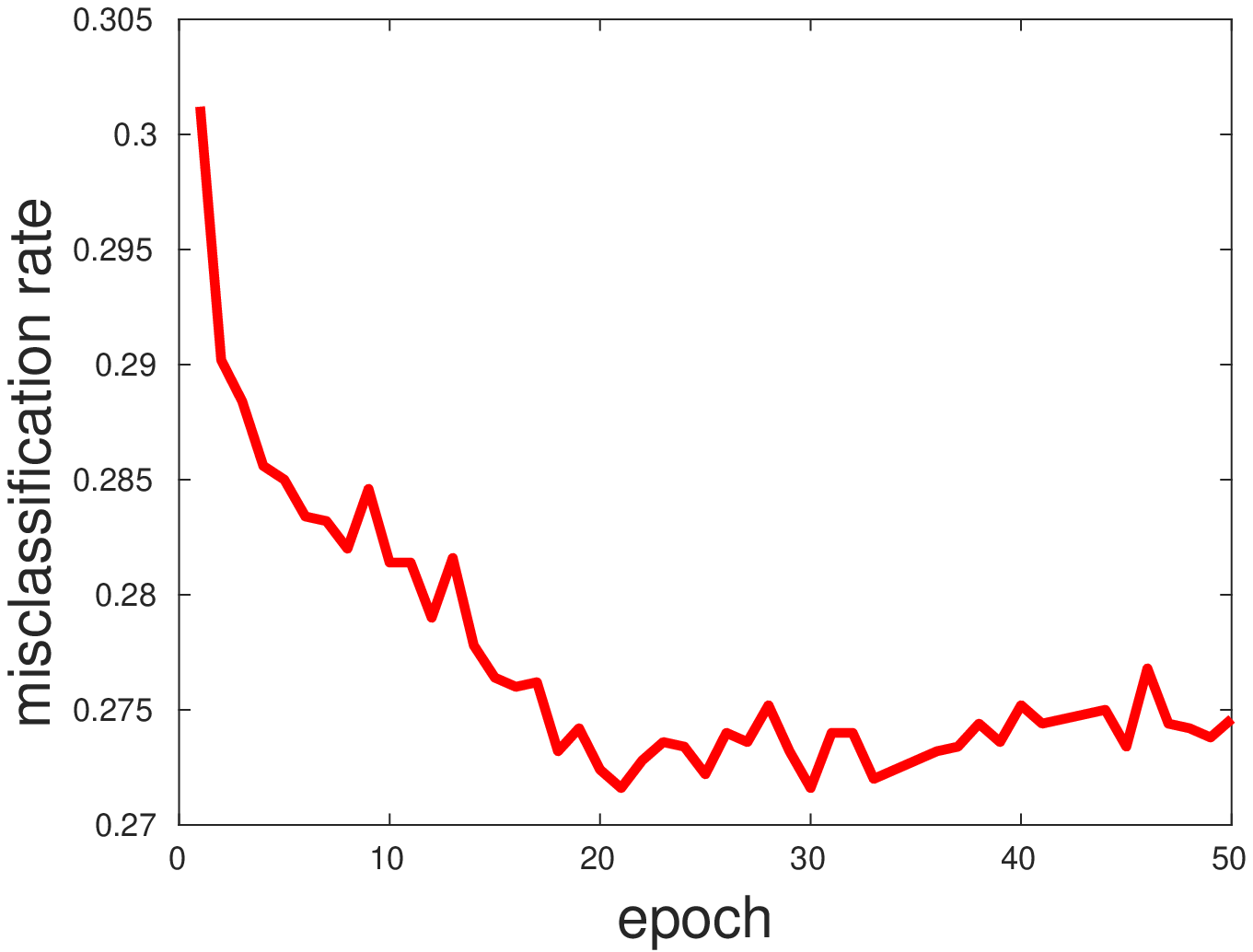}
                    \caption{Misclassification rate.}
                    \label{cdm_accuracy}
              \end{subfigure}
              \caption{ (a) AMI fitting score of clusters via cross-validation based on fully random initialization. Converged clusters tend to match with the other. (b) The misclassification rate decreases as training proceeds and the prediction accuracy converges to a comparable value with GS1MC even CDMC did not take any preliminary cluster information.}
        \end{figure}

        To start with, we split the data (95\%, 5\%) and initialize group identities of users/items randomly. Then we train the \textit{CDMC} model for a number of epochs until the clustering results tend to stabilize (200 epochs for 95\% Movielens $100k$ dataset). So far, the produced cluster identities of each instance are stored as the baseline. Afterwards, we re-conduct the process multiple times with completely random initialization. Namely, another (95\%, 5\%) entries of the dataset are split for cross-validation, and all the cluster identities, as well as all latent variables, are determined arbitrarily. We use adjusted mutual information (AMI) \cite{vinh2010information} as the evaluation score to measure the degree of matching, regarding the clustering results from multiple cross-validation processes. As Figure \ref{AMI_cluster} shows, both user/item clusters converge to a highly similar distribution over the training epochs.

        Meanwhile, during each iteration of the optimization process, we construct $f(\mathbf{M})$ and make the prediction on the $5\%$ test set. The recorded misclassification rate is shown in Figure \ref{cdm_accuracy}. It is indicated that the misclassification rate gradually stabilize as the cluster developing process proceeds and the resulted prediction accuracy is highly comparable with GS1MC proposed in Session \ref{Sec:3}, even for this case the cluster information is totally unknown. 
        
        \subsubsection{Clustering Outcome}
\begin{figure}
              \begin{subfigure}[b]{0.48\textwidth}
                    \includegraphics[width=\textwidth]{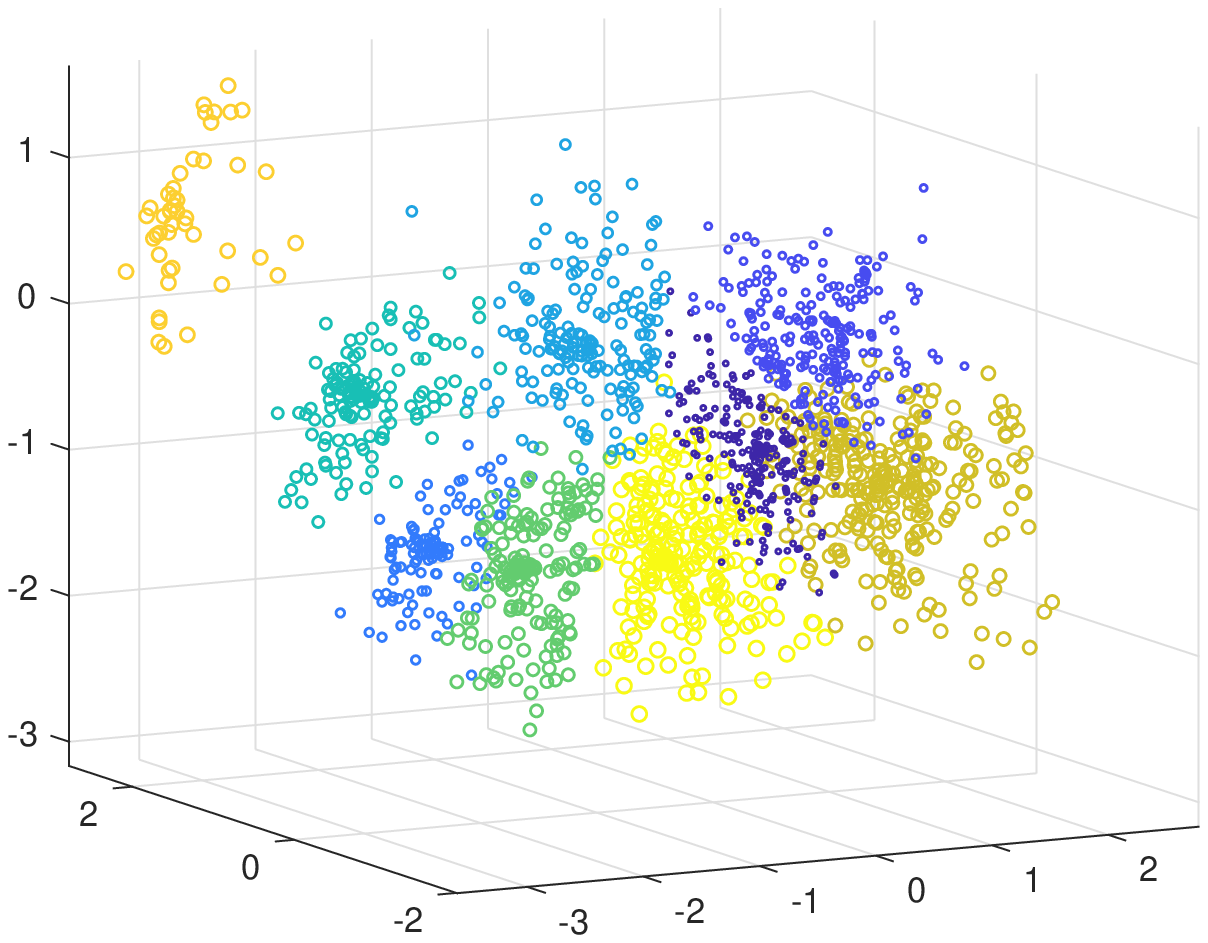}
                    \caption{Item clusters.}
                    \label{item_ssc}
              \end{subfigure}
              \begin{subfigure}[b]{0.48\textwidth}
                    \includegraphics[width=\textwidth]{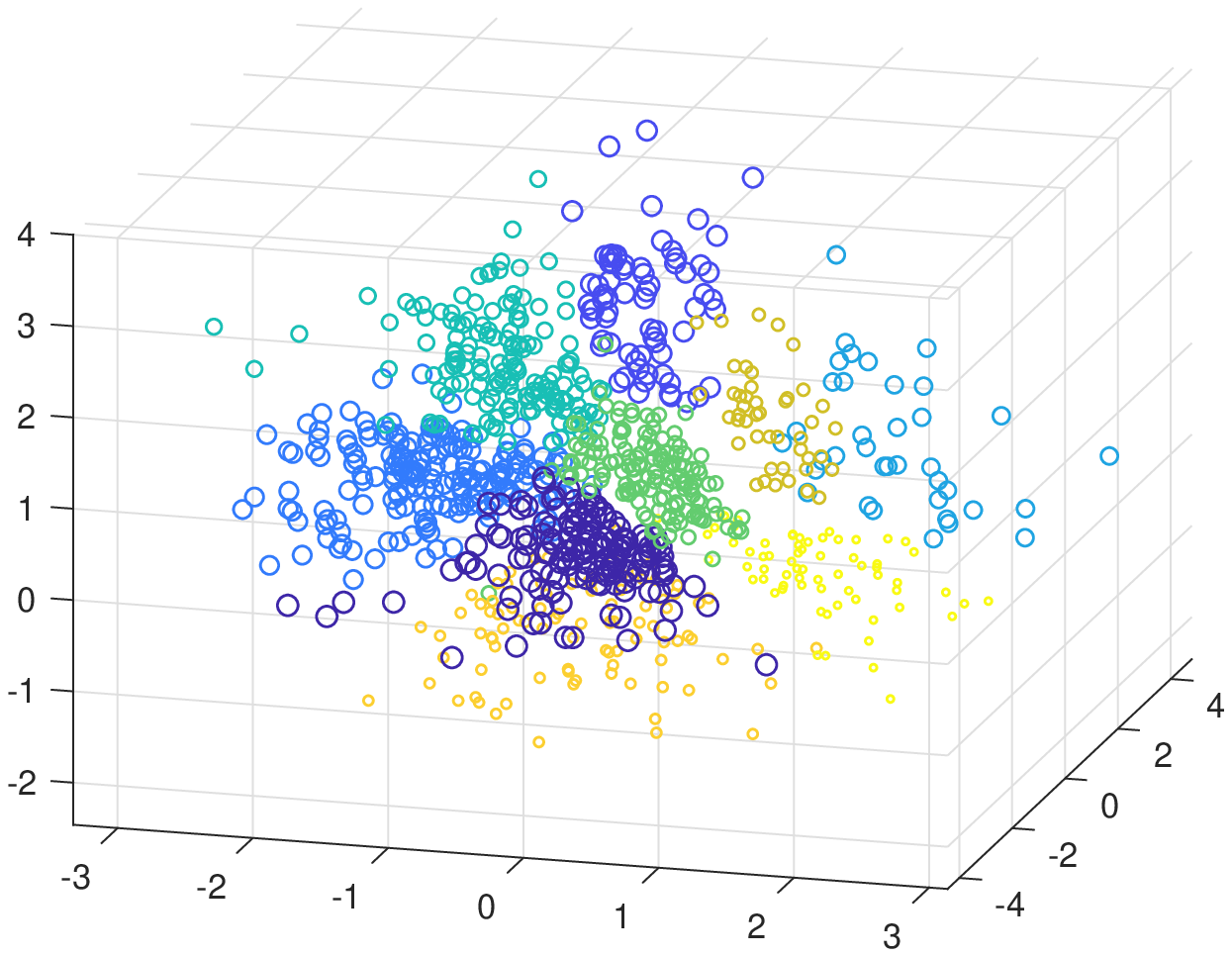}
                    \caption{User clusters.}
                    \label{user_ssc}
              \end{subfigure}
            \caption{The clustering results of CDMC.}
            \label{ssc_cluster}
        \end{figure}

        For item clusters, we use three dimensions of item-related latent variable $(\mathbf{Q}+\mathbf{I}^T_J\mathbf{T}_J)$ as axis. The learned clusters are visualized in Figure \ref{item_ssc}. Similarly, developed user clusters are plotted on user-related latent variable $(\mathbf{P}+\mathbf{I}^T_U\mathbf{S}_U)$. As Figure \ref{ssc_cluster} shows, the item clusters are more dispersed and differentiable while the user clusters gather in closer proximity. 
        
        In order to validate the practical influence of CDMC, we project the actual profile features of each user/item onto the latent variable CDMC learned and discovered some noteworthy findings. 

        Firstly, since there are 19 categories of items available, and each movie can be labeled as multiple genres. We extracted this information and constructed a genre matrix $\mathbf{A} \in \mathbb{R}^{n_2 \times 19}$, here $a_{ig} = 1$ means item-$i$ can be classified in category-$g$. As items in $\mathbf{A}$ share the 1-to-1 exact same index with $(\mathbf{Q}+\mathbf{I}^T_J\mathbf{T}_J)$, we applied k-means clustering method on this generic information and visualized its results corresponding to the latent variable that CDMC learned. 
        
\begin{figure}
            \begin{subfigure}[b]{0.32\textwidth}
                \includegraphics[width=\textwidth]{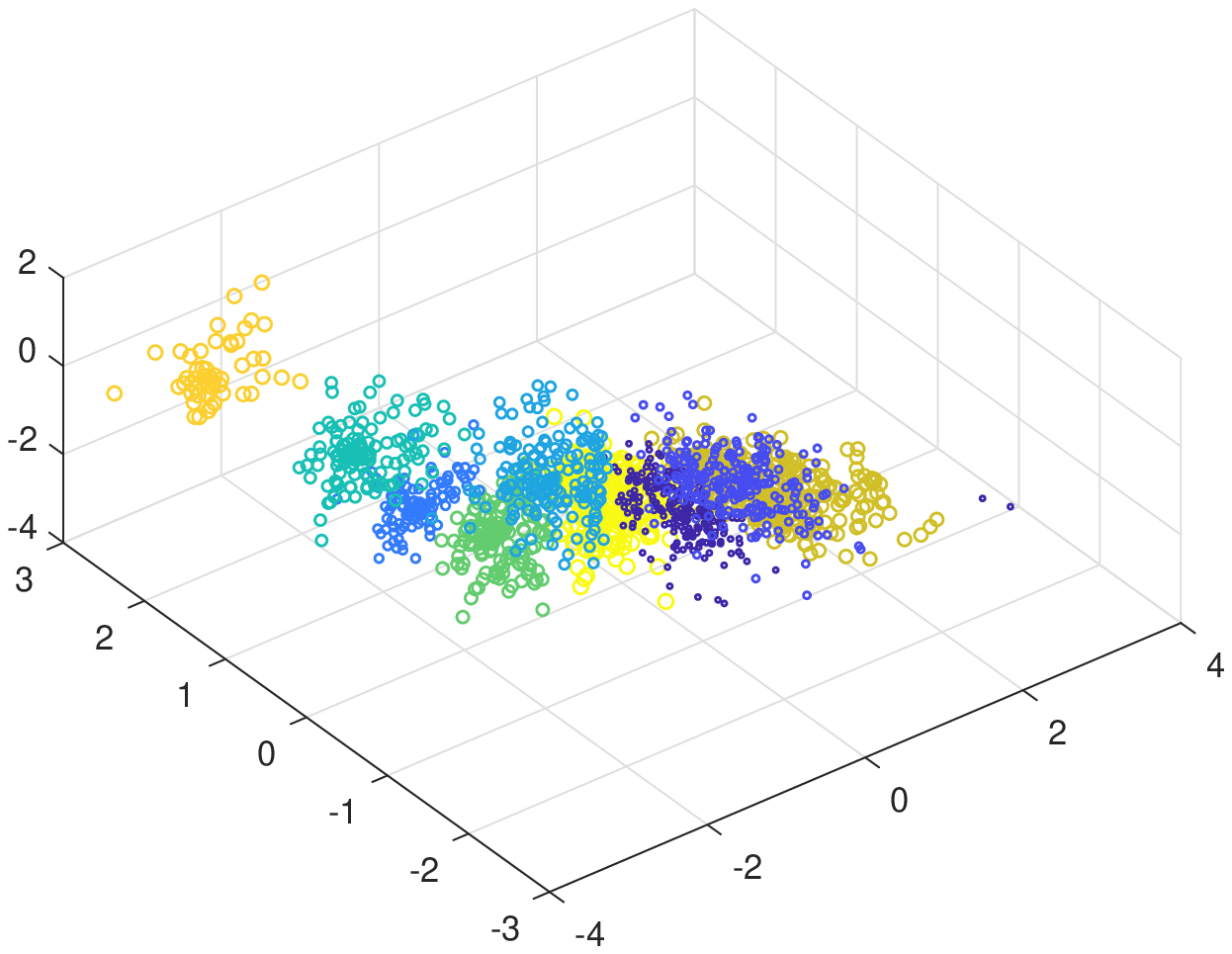}
                \caption{CDMC: Items.}
                \label{item_ssc_genre}
          \end{subfigure}
          \begin{subfigure}[b]{0.32\textwidth}
                \includegraphics[width=\textwidth]{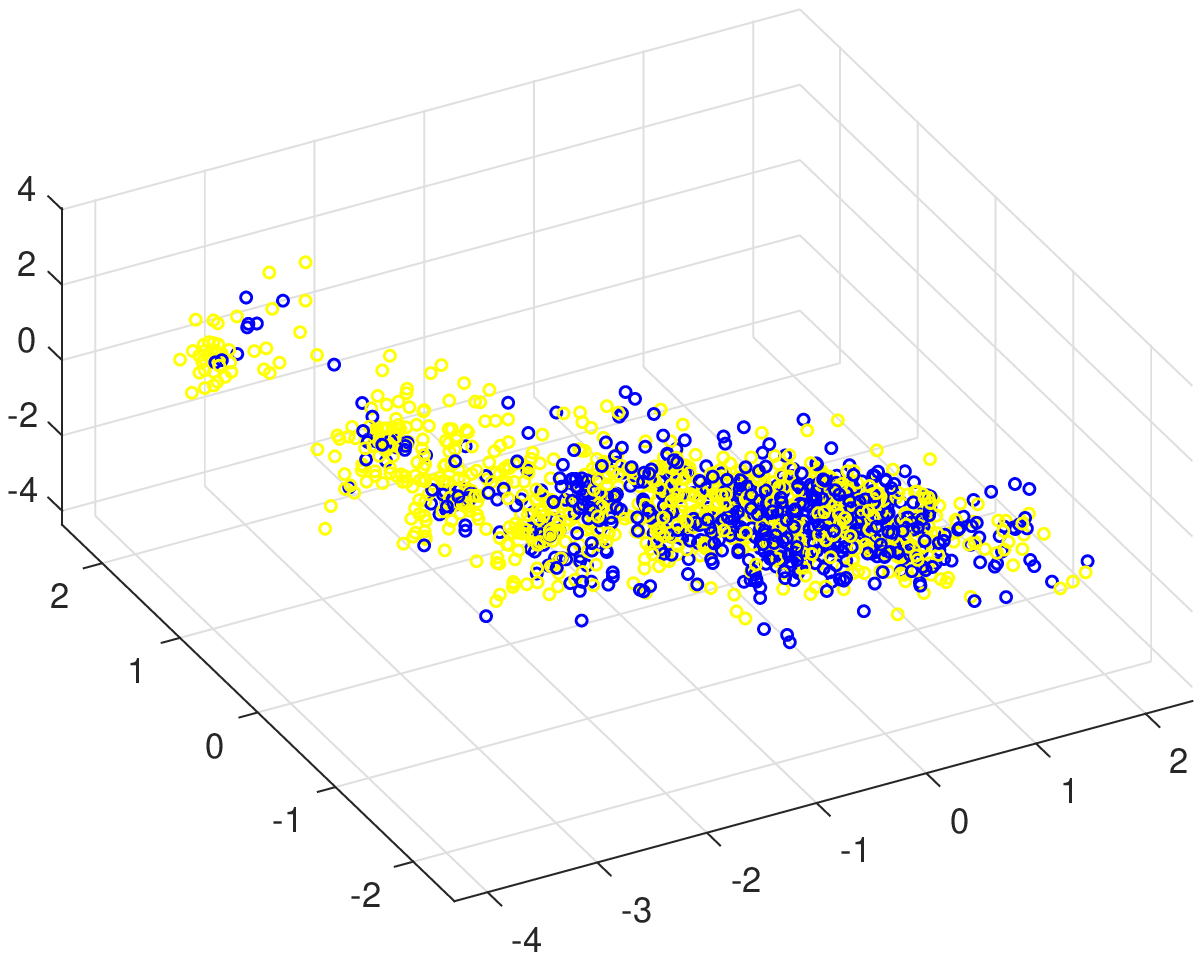}
                \caption{Item Genres.}
                \label{kmeans_genre}
          \end{subfigure}
          \begin{subfigure}[b]{0.32\textwidth}
                \includegraphics[width=\textwidth]{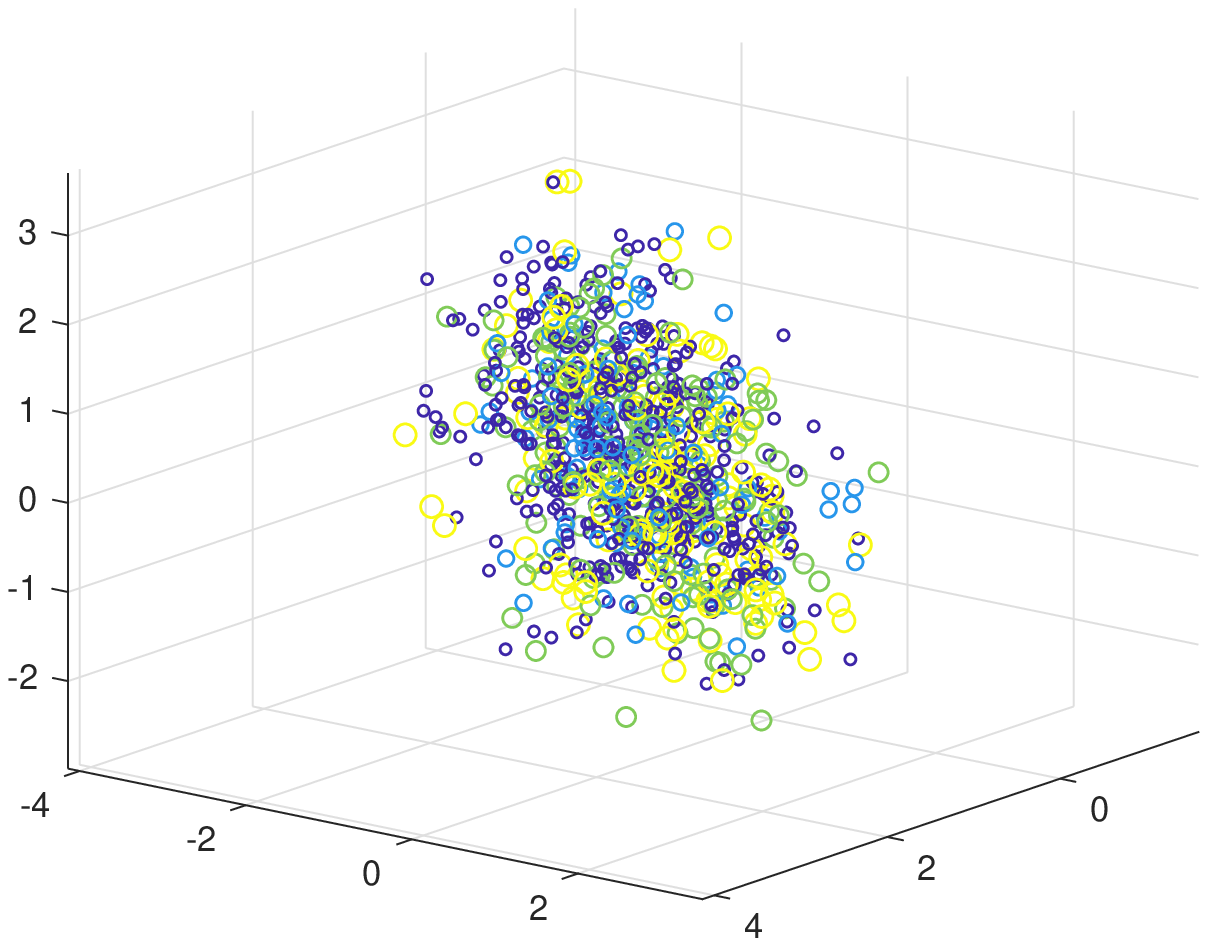}
                \caption{User profile.}
                \label{kmeans_profile}
          \end{subfigure}
            \caption{(a) Item clustering results of CDMC. (b) Items' genre information is reflected on the latent variable in a clear differentiable pattern. (c) User clusters show much higher complexity, but it is still noticeable that blue and purple nodes  aggregate on vertically higher part of the space while yellow and green points are distributed below.}
        \end{figure}

        As shown in Figure \ref{kmeans_genre}, it is compelling that the learned latent variable $(\mathbf{Q}+\mathbf{I}^T_J\mathbf{T}_J)$ have a clear discernible pattern regarding items' generic features. In other words, even though the fact that our proposed CDMC method did not take any generic information, it has captured items' factual profile based on only the sparse rating matrix. Besides, as CDMC conducts sparse subspace clustering and group-specific matrix completion in an iterative manner, along with gradually learning the hidden profiles, the model can integrate this information immediately into matrix completion task, which in turn positively boost the next iteration's clustering.
        
        Similarly, we build a feature matrix of users based on their context profiles, including \textit{age, gender} and \textit{occupations}. The clustering result is projected on $(\mathbf{P}+\mathbf{I}^T_U\mathbf{S}_U)$ and shown in Figure \ref{kmeans_profile}. As we expected, understanding human's preference is a much more complicated task, and the clustering result is visibly more confusing. But it is still noticeable in the plot that blue and purple nodes gather in the vertically higher part of the space while yellow and green ones are distributed below. As pointed out in the previous literature \cite{2012arXiv1208.1544Z}, it is a quite common issue that multiple individuals might share a single account, which has biased the accuracy of the profile information. 

\section{Conclusions and Future Works}\label{Sec:6}
    In this paper, we introduced group-specific matrix factorization into 1-bit matrix completion task and proposed GS1MC. Then we first time integrated sparse subspace clustering with matrix completion task and proposed CDMC, extending the scope of GS1MC from passively receiving preliminary cluster information into positively developing clusters and leveraging their effects. Experiments show GS1MC outperforms existing methods on both synthetic and real-world data, especially for the \textit{cold-start problem}. And CDMC successfully captures items' hidden generic features from highly sparse binary rating matrix. It is noteworthy that GS1MC and CDMC provide a new insight to evaluate the quality of clusters or to detect undiscovered segments. For instance, when integrating implicit feedback clusters into GS1MC, the prediction accuracy was greatly boosted compared to previous methods. In terms of  CDMC, our experiments show movies' genres have a large impact on their popularity among certain audience while users' age, gender and occupation tends to have slighter effects on their preference. For future work, it will be valuable to apply GS1MC and CDMC into more real-world applications and discover possible unrevealed social behavior and market phenomenon.

\bibliographystyle{splncs04}

\end{document}